\documentclass[twocolumn]{cinc}

\usepackage{graphicx}
\usepackage{hyperref}
\usepackage{amsmath}

\begin{document}

\bibliographystyle{cinc}
\graphicspath{{images/}{../images/}}

%%%%%%%%%%%%%%%%%%%%%%%%%%%%%%%%%%%%%%%%%%%%%%%%%
\title{Combining Hough Transform and Deep Learning Approaches\\ to Reconstruct ECG Signals From Printouts}

\author{Felix Krones$^{1}$, Ben Walker$^{2}$, Terry Lyons$^{2}$, Adam Mahdi$^{1}$ \\
\ \\
$^1$ Oxford Internet Institute, University of Oxford, UK\\
$^2$ Mathematical Institute, University of Oxford, UK}

\maketitle
%%%%%%%%%%%%%%%%%%%%%%%%%%%%%%%%%%%%%%%%%%%%%%%%%

%%%%%%%%%%%%%%%%%%%%%%%%%%%%%%%%%%%%%%%%%%%%%%%%%
\begin{abstract}

    % Background
    This work presents our team's (SignalSavants) winning contribution to the 2024 George B. Moody PhysioNet Challenge. The Challenge had two goals: reconstruct ECG signals from printouts and classify them for cardiac diseases. Our focus was the first task.
    Despite many ECGs being digitally recorded today, paper ECGs remain common throughout the world. Digitising them could help build more diverse datasets and enable automated analyses. However, the presence of varying recording standards and poor image quality requires a data-centric approach for developing robust models that can generalise effectively.
    % Experiments
    Our approach combines the creation of a diverse training set, Hough transform to rotate images, a U-Net based segmentation model to identify individual signals, and mask vectorisation to reconstruct the signals.
    We assessed the performance of our models using the 10-fold stratified cross-validation (CV) split of 21,799 recordings proposed by the PTB-XL dataset. 
    % Results
    On the digitisation task, our model achieved an average CV signal-to-noise ratio of $17.02$ and an official Challenge score of $12.15$ on the hidden set, securing first place in the competition.
    % Conclusion/Implication
    Our study shows the challenges of building robust, generalisable, digitisation approaches. Such models require large amounts of resources (data, time, and computational power) but have great potential in diversifying the data available.

\end{abstract}
%%%%%%%%%%%%%%%%%%%%%%%%%%%%%%%%%%%%%%%%%%%%%%%%%

%%%%%%%%%%%%%%%%%%%%%%%%%%%%%%%%%%%%%%%%%%%%%%%%%
\section{Introduction}
%%%%%%%%%%%%%%%%%%%%%%%%%%%%%%%%%%%%%%%%%%%%%%%%%
% Background
Cardiovascular diseases remain the leading cause of death globally \cite{whoCardiovascularDiseases}, with electrocardiograms (ECGs) being essential for diagnosis and monitoring. Despite the rise of digital ECG devices, many ECGs are still recorded on paper, particularly in resource-limited settings \cite{challenge2024}. Digitising these recordings could enhance datasets with greater demographic and temporal diversity, but the task is complicated by varying recording standards and image qualities.

% Knowledge gap & Hypothesis/purpose/goal
Although digitisation efforts have advanced, they often focus on a narrow set of images, leading to models that may not generalise well across different populations and recording conditions. The 2024 George B. Moody PhysioNet Challenge \cite{challenge2024} addresses this gap by calling for robust algorithms to reconstruct ECG signals from printouts.

% Approach/Contribution
The 2024 Challenge included two tasks: recovering ECG signals from printouts and predicting cardiac diseases from these signals. Our team, {\it SignalSavants}, concentrated on the first task, with the goal of creating a versatile approach that can be trained and tested on diverse image sources by combining traditional image processing with deep learning techniques. We implemented an adaptable pipeline that includes image rotation using the Hough line transform, signal segmentation using deep learning, and mask vectorisation. We rigorously validated our models using cross-validation splits of 21,799 recordings from the PTB-XL dataset \cite{wagner2022ptbxl, wagner2020ptbxl}, while creating multiple image versions for each signal to ensure robustness and reliability.

%%%%%%%%%%%%%%%%%%%%%%%%%%%%%%%%%%%%%%%%%%%%%%%%%
\section{Methodology}
%%%%%%%%%%%%%%%%%%%%%%%%%%%%%%%%%%%%%%%%%%%%%%%%%
%------------------------------------------
\subsection{Data}
\begin{figure}[htbp]
\centering
    \includegraphics[width=\linewidth]{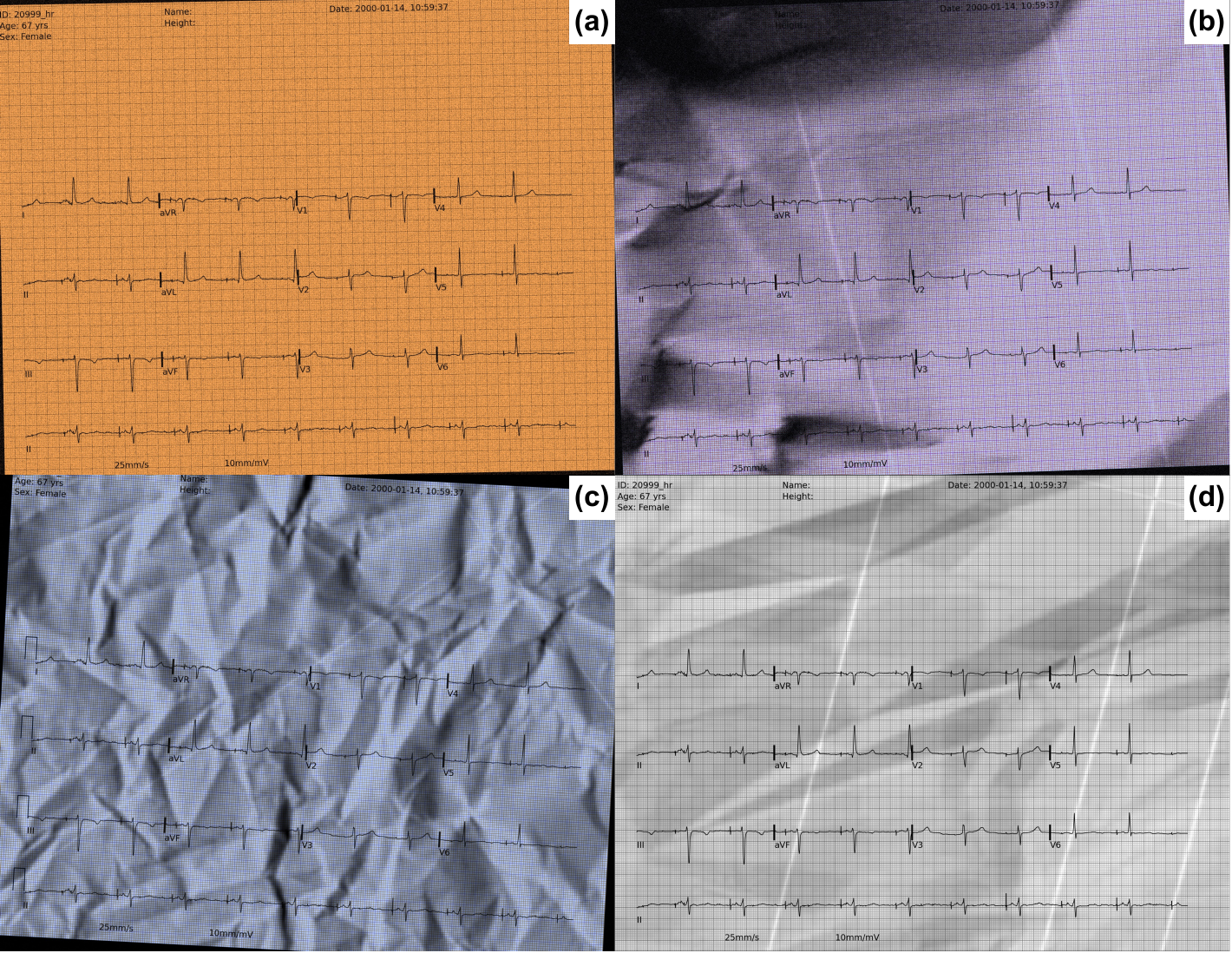}
    \caption{Four example images of signal 20999. (a) rotated by one degree, (b) rotated by two degrees with wrinkles and shadows, (c) rotated by three degrees with wrinkles and shadows, (d) not rotated with wrinkles and shadows.}
    \label{fig:examples}
\end{figure}

The PTB-XL dataset \cite{wagner2022ptbxl, goldberger2000physiobank} contains 21,799 12-lead ECG recordings from 18,869 patients. The data was collected from October 1989 to June 1996 and publicly released in 2019.
For each of the recordings, we created at least four images using randomly applied augmentation options from the \texttt{ecg-image-kit} \cite{Shivashankara2024, ECGImageKit2024}. We used a maximum rotation of 30 degrees and default settings for the other augmentation options. Figure \ref{fig:examples} provides examples. Table \ref{tab:scores} describes the randomly used augmentation options.
We assumed the following standards across images, which we discuss further below: the same signal order, the same relative signal position, the same grid scale and the same amount of full-length signals.

%------------------------------------------
\subsection{Architecture}\label{sec:models}
\begin{figure}[htbp]
\centering
    \includegraphics[width=\linewidth]{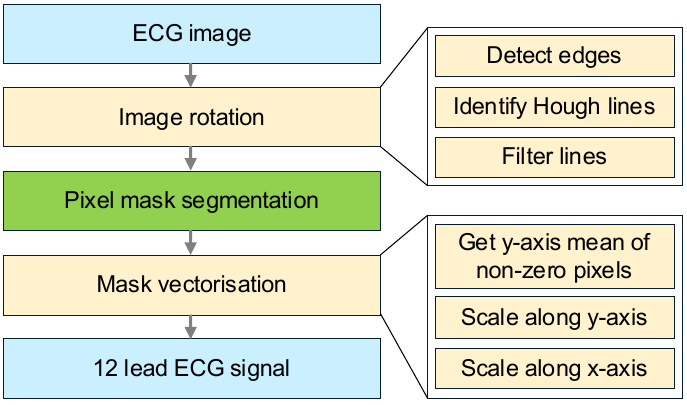}
    \caption{A schematic diagram of our model architecture. Blue: Data, Yellow: Rule-based engineering, Green: Deep learning.}
    \label{fig:architecture}
\end{figure}

Our approach consisted of three major steps: image rotation, segmenting the image into individual signals, and vectorising the segmented pixels. Figure \ref{fig:architecture} is a schematic diagram outlining this approach. 
%We first rotated and pre-processed all images. Next, we used a deep learning model to segment the masks along with their labels from the images. Finally, we vectorised and scaled the segmented pixels.

%------------------------------------------
\medskip\noindent
{\bf Rotation.}
We rotated the ECG images to standardise the subsequent tasks. To achieve this, we used the \texttt{HoughLines()} method from the \texttt{opencv} (\texttt{cv2}) Python package. This method detects straight lines in an image by mapping edge points onto a polar coordinate system and finding intersections that indicate the presence of a line. We applied filters to these detected lines, considering only those within a specific angle range and ensuring a minimum number of parallel lines. Using the angles of these filtered lines, we calculated the rotation to align the ECG waveforms horizontally.
%Figure \ref{fig:rotation} provides an example of the filtered lines.

% \begin{figure}[htbp]
% \centering\includegraphics[width=\linewidth]{images/rotation.png}
%     \caption{Example of all (left) and filtered (right) Hough lines (in blue) used to determine rotation for signal 20999.}
%     \label{fig:rotation}
% \end{figure}

%------------------------------------------
\medskip\noindent
{\bf Segmentation.}\label{method:seg}
For the segmentation we used nnU-Net \cite{isensee2021nnu}, which was developed by the German Cancer Research Center. nnU-Net automatically configures the image processing and network architecture based on the data, using three types of parameters. First, it uses a set of fixed parameters (which remain constant due to their robust nature), such as the loss function, primary data augmentation strategies and learning rate. Second, rule-based parameters are adapted to the dataset using predefined heuristic rules, adjusting aspects like network topology and batch size. Finally, empirical parameters are determined through trial-and-error, refining elements like the best U-net configuration for the dataset and the postprocessing strategy.

\begin{figure}[htbp]
\centering
    \includegraphics[width=\linewidth]{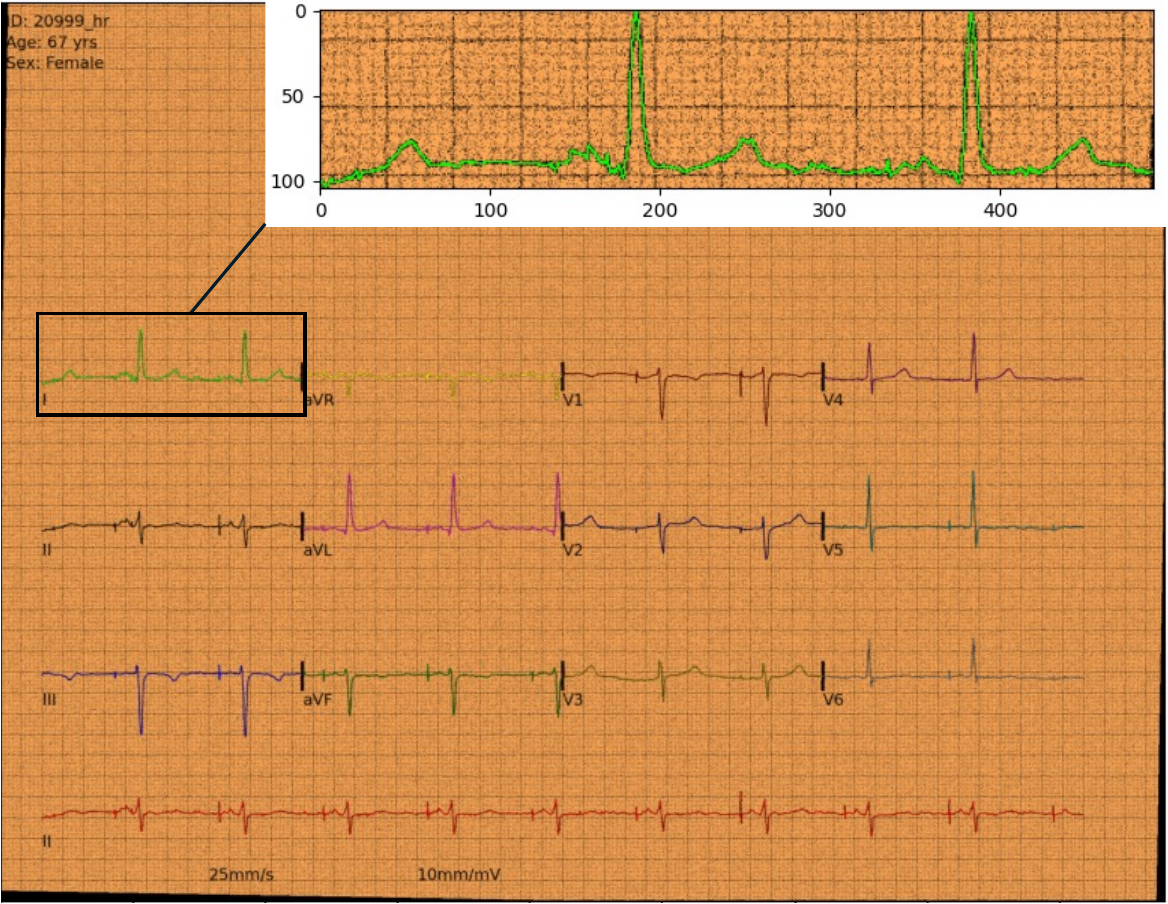}
    \caption{Example of predicted masks for signal 20999.}
    \label{fig:segmentation}
\end{figure}

Figure \ref{fig:segmentation} shows a segmentation example.
To create the masks we used two sets of pixels. The ones provided by the \texttt{ecg-image-kit} (called `sparse') and an interpolated version that increased the pixel density (called `dense').

\begin{figure}[htbp]
\centering\includegraphics[width=\linewidth]{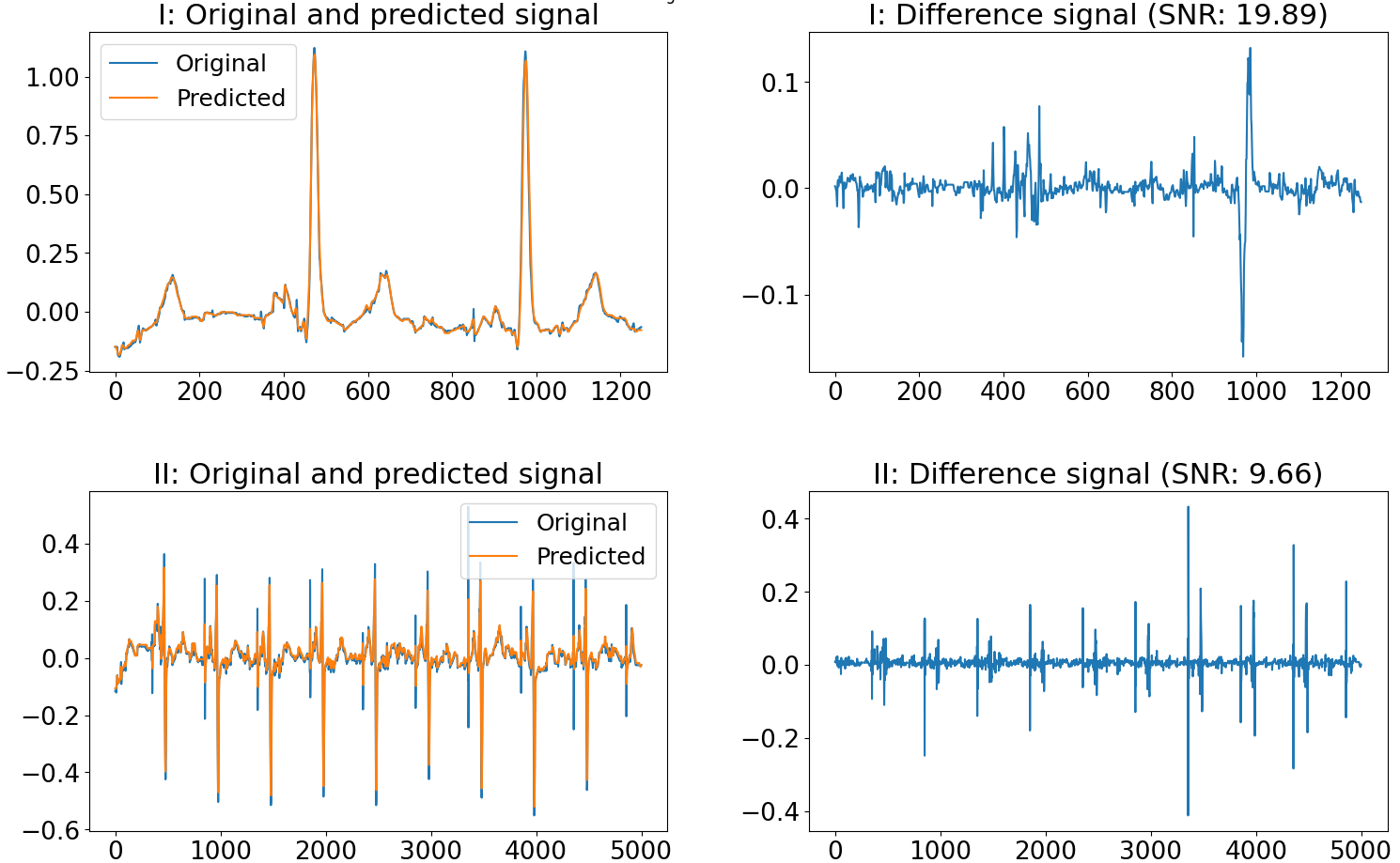}
    \caption{Vectorisation example for signal 20999. {\it Left}: original and predicted signal. {\it Right}: signal difference.}
    \label{fig:vectorisation}
\end{figure}

%------------------------------------------
\medskip\noindent
{\bf Vectorisation.} 
In order to convert the mask, which is a matrix of pixel values, into a signal, the average y-position of the non-zero pixels in each column is used as signal value. Next, the signal is scaled to the correct amplitude using the grid size and resolution, which we assume to be constant. Finally, the starting time of the signal is determined based on the position of the signal on the ECG. \autoref{fig:vectorisation} shows an example of two leads from an ECG and the predicted signals from vectorising a mask of the signals.

%------------------------------------------
\subsection{Scoring}
The metric used in the the digitisation task was the signal-to-noise ratio (SNR) between the true signal \( y \) and the predicted signal \( \hat{y} \), which can be calculated via
\begin{equation}\label{eq:accu}
\text{SNR} = 10 \cdot \log_{10} \frac{\sum_i y_i^2}{\sum_i (\hat{y}_i - y_i)^2}.
\end{equation}
Letting the prediction be $\hat{y}_i=0$ for all $i$ provides a surprisingly strong baseline of $\text{SNR}=0$ for this Challenge. This is due to the ECG signals having high and narrow peaks. During the Challenge, an update was released that aligns predicted and true signals up to a certain threshold along the $x$ and $y$-axis \cite{challenge2024}.

To evaluate our models, we used the 10-fold splits suggested in \cite{wagner2022ptbxl} due to their stratified sampling approach.

%%%%%%%%%%%%%%%%%%%%%%%%%%%%%%%%%%%%%%%%%%%%%%%%%
\section{Results}
%%%%%%%%%%%%%%%%%%%%%%%%%%%%%%%%%%%%%%%%%%%%%%%%%
Table \ref{tab:rank} shows the final challenge scores and Table \ref{tab:scores} the results of different models for the digitisation task. Models M2 and M3, while in theory more advanced, were more prone to overfitting and needed more training data to achieve similar performance as M1.
In the rotation step, we achieved a correct rotation prediction on over 99.7\% of the images. Manually investigating a selection of the images where the rotation failed suggests that failure occurs when the colour schema and/or the augmentation are so extreme that determining the straight lines becomes very difficult. Vectorisation sometimes did not capture the peaks of the signals, as illustrated in \autoref{fig:vectorisation}.

\begin{table}[!ht]
    \centering
   \begin{tabular}{llr}
    \hline\hline
    Task & Score & Rank \\
        \hline
            Digitization    & SNR: 12.15  & 1/16 \\
            Classification & F-measure: - & - \\
        \hline\hline
    \end{tabular}
    \caption{Signal-to-noise (SNR) ratio and F-measure of our team’s model on the hidden data for the digitization and classification (not participated) tasks, respectively.}
    \label{tab:rank}
\end{table}

\begin{table*}[!ht]
    \centering
   \begin{tabular}{clllll}
    \hline\hline
    Model & Train images & Augmentation & Mask & SNR official & SNR CV \\
        \hline
            M1  & 20k   & Headers, calibration, wrinkles, rotation, cropping & Sparse     & 12.15    & 17.02 \\
            M2  & 45k   & M1 + handwritten notes, QR-codes, various colours & Dense     &  9.42     & 12.58   \\
            M3  & 69k + 100 fine & Same as M2 & Dense     &  4.20     & 15.27   \\
        \hline\hline
    \end{tabular}
    \caption{Digitisation models tested in this work. `SNR official' is the leaderboard performance on the Challenge’s hidden set. `SNR CV' is the average CV score. The column `Augmentation' describes the randomly applied options from the \texttt{ecg-image-kit} during image generation. `Dense' means an interpolated set of pixels was used for the masks (see \ref{method:seg}). M3 was first trained locally on 69k images and then further fine-tuned on 100 of the hidden Challenge train data.}
    \label{tab:scores}
\end{table*}

In addition to the PTB-XL scans used in the official ranking, models were evaluated on a variety of additional image types \cite{challenge2024data}. Although we still achieved positive scores (0.51 to 4.93) on both colour and black-and-white scans of various paper qualities (deteriorated and clean), we scored negatively (-1.76 to -0.72) on mobile phone photos and computer screenshots.

%%%%%%%%%%%%%%%%%%%%%%%%%%%%%%%%%%%%%%%%%%%%%%%%%
\section{Discussion}
%%%%%%%%%%%%%%%%%%%%%%%%%%%%%%%%%%%%%%%%%%%%%%%%%
In this study, we combined several common image processing approaches, such as the Hough transform for line detection, U-Nets for segmentation, and interpolation techniques for scaling. This method achieved the highest Challenge score, surpassing the next best-performing team by more than double.

However, the leaderboard only shows results for scans from the PTB-XL dataset, while the Challenge also evaluated entries on scans and mobile phone photos of images of different quality (e.g., clean, stained and deteriorated) as well as screenshots of computer monitors. No team achieved the highest score across all datasets. In fact, leaderboard performance dropped significantly on the other datasets (e.g., no team received a positive score for mobile phone photos of stained papers). Additionally, our local CV scores on various data augmentations were much higher than on the hidden sets. This performance gap highlights the challenge of achieving robust generalisation and the need for a training set that closely matches real-world conditions for reliable digitisation.

Some of our assumptions affected our results. Notably, all training examples had the signals in the same order and at the same position. While the segmentation model has the ability to learn the labels from the labels in the images, we assume that it simply assumed the same signal position for each prediction. We tested combining bounding boxes and optical character recognition (OCR) to be more independent of the signal position, but did not achieve robust results locally. Additionally, we assumed that each image contains only one full signal, which may not always be the case. For better generalisation, the segmentation model must be trained on examples with multiple full signals. We also assumed constant image resolution and grid scaling. 

%One of the most challenging tasks was to achieve the correct y-scaling without relying on image assumptions, since the original evaluation score compared absolute values, not just relative changes in signals.

%\textit{Related work.}
Our approach to ECG digitisation aligns with the multi-step approaches commonly reported in the literature \cite{fortune2021digitizing, baydoun2019high, wu2022fully, li2020deep}. Initially, we applied image pre-processing using the Hough transform \cite{baydoun2019high, wu2022fully}. While many studies relied on pixel thresholds for segmentation \cite{baydoun2019high, fortune2021digitizing, wu2022fully}, our objective was to improve generalisability by using a deep learning network (nnU-Net \cite{isensee2021nnu}), similar to the approach in \cite{li2020deep}, followed by vectorisation. However, unlike previous work \cite{li2020deep}, we applied our model to the entire image without prior pre-processing (except rotation), such as grid removal or detection of regions of interest. The key contributions of this approach are that a) it achieves much better results than other submissions to digitise ECG signals, and b) it can be easily applied to different data sources, whereas previous approaches are highly dependent on the specific style of the images used (e.g., assuming a particular grid colour). As long as the training data mimic the deployment settings, our approach can be easily adjusted.

%\textit{Conclusion/Big picture.}
%Separating the steps of data pre-processing and prediction modelling can be beneficial for researchers, as it allows them to focus on model improvement and facilitates the sharing of downstream models across different settings.

Developing methods to digitise images can enhance machine learning-supported screening systems by improving training through a wider range of data and helping to apply the models in more settings.
However, making models robust and generalisable is not trivial. Achieving this requires a data-centric approach and vast resources.

% %%%%%%%%%%%%%%%%%%%%%%%%%%%%%%%%%%%%%%%%%%%%%%%%%
% \section*{Code availability}
% %%%%%%%%%%%%%%%%%%%%%%%%%%%%%%%%%%%%%%%%%%%%%%%%%
% Our complete code is available on \href{https://github.com/felixkrones/physionet24}{GitHub}.

%%%%%%%%%%%%%%%%%%%%%%%%%%%%%%%%%%%%%%%%%%%%%%%%%
\section*{Acknowledgements}  
%%%%%%%%%%%%%%%%%%%%%%%%%%%%%%%%%%%%%%%%%%%%%%%%%
We thank the Applied Computer Vision Lab of the Helmholtz Institute and the Medical Image Computing Devision at the DKFZ for making nnU-Net publicly available. FK was partially supported by the German tax payer through the Friedrich Naumann Foundation and BW and TL by the Hong Kong Innovation and Technology Commission and the Centre for Intelligent Multidimensional Data Analysis. TL was funded in part by the EPSRC [EP/S026347/1], The Alan Turing Institute [EP/N510129/1], the Defence and Security Programme, and the Office for National Statistics.

%%%%%%%%%%%%%%%%%%%%%%%%%%%%%%%%%%%%%%%%%%%%%%%%%
\bibliography{refs}

\begin{thebibliography}{10}
\expandafter\ifx\csname url\endcsname\relax
  \def\url#1{\texttt{#1}}\fi
\expandafter\ifx\csname urlprefix\endcsname\relax\def\urlprefix{URL }\fi

\bibitem{whoCardiovascularDiseases}
{World Health Organisation}.
\newblock {Cardiovascular diseases (CVDs)}.
\newblock Geneva; 2021 Jun 11 [cited 2023 May 10]. Available from: \url{https://www.who.int/en/news-room/fact-sheets/detail/cardiovascular-diseases-(cvds)}.

\bibitem{challenge2024}
Reyna MA, Deepanshi, Weigle J, Koscova Z, Elola A, Seyedi S, Campbell K, Clifford GD, Sameni R.
\newblock {Digitization and Classification of ECG Images: The George B. Moody PhysioNet Challenge 2024} ;\hspace{0pt}51:1--4.

\bibitem{wagner2022ptbxl}
Wagner P, Strodthoff N, Bousseljot R, Samek W, Schaeffter T.
\newblock {PTB-XL, a large publicly available electrocardiography dataset (version 1.0.3)}.
\newblock \url{https://doi.org/10.13026/kfzx-aw45}, 2022.

\bibitem{wagner2020ptbxl}
Wagner P, Strodthoff N, Bousseljot RD, Kreiseler D, Lunze F, Samek W, Schaeffter T.
\newblock {PTB-XL: A Large Publicly Available ECG Dataset}.
\newblock Scientific Data 2020;\hspace{0pt}.

\bibitem{goldberger2000physiobank}
Goldberger AL, Amaral LA, Glass L, Hausdorff JM, Ivanov PC, Mark RG, Mietus JE, Moody GB, Peng CK, Stanley HE.
\newblock {PhysioBank, PhysioToolkit, and PhysioNet: Components of a new research resource for complex physiologic signals}.
\newblock Circulation ;\hspace{0pt}101(23):e215--e220.

\bibitem{Shivashankara2024}
Shivashankara KK, Deepanshi, Shervedani AM, Reyna MA, Clifford GD, Sameni R.
\newblock {ECG-Image-Kit: a synthetic image generation toolbox to facilitate deep learning-based electrocardiogram digitization}.
\newblock Physiological Measurement 2024;\hspace{0pt}45:055019.

\bibitem{ECGImageKit2024}
Deepanshi, Shivashankara K, Clifford GD, Reyna MA, Sameni R.
\newblock {ECG-Image-Kit: A Toolkit for Synthesis, Analysis, and Digitization of Electrocardiogram Images}, January 2024.

\bibitem{isensee2021nnu}
Isensee F, Jaeger PF, Kohl SA, Petersen J, Maier-Hein KH.
\newblock {nnU-Net: a self-configuring method for deep learning-based biomedical image segmentation}.
\newblock Nature Methods 2021;\hspace{0pt}18(2):203--211.

\bibitem{challenge2024data}
Reyna MA, Deepanshi, Weigle J, Koscova Z, Campbell K, Shivashankara KK, Saghafi S, Nikookar S, Motie-Shirazi M, Kiarashi Y, Seyedi S, Clifford GD, Sameni R.
\newblock {ECG-Image-Database: A Dataset of ECG Images with Real-World Imaging and Scanning Artifacts; A Foundation for Computerized ECG Image Digitization and Analysis}, 2024.
\newblock \urlprefix\url{https://arxiv.org/abs/2409.16612}.

\bibitem{fortune2021digitizing}
Fortune JD, Coppa NE, Haq KT, Patel H, Tereshchenko LG.
\newblock {Digitizing ECG image: new fully automated method and open-source software code}.
\newblock medRxiv 2021;\hspace{0pt}2021--07.

\bibitem{baydoun2019high}
Baydoun M, Safatly L, Abou~Hassan OK, Ghaziri H, El~Hajj A, Isma’eel H.
\newblock High precision digitization of paper-based ecg records: a step toward machine learning.
\newblock IEEE Journal of Translational Engineering in Health and Medicine 2019;\hspace{0pt}7:1--8.

\bibitem{wu2022fully}
Wu H, Patel KHK, Li X, Zhang B, Galazis C, Bajaj N, Sau A, Shi X, Sun L, Tao Y, et~al.
\newblock {A fully-automated paper ECG digitisation algorithm using deep learning}.
\newblock Scientific Reports 2022;\hspace{0pt}12(1):20963.

\bibitem{li2020deep}
Li Y, Qu Q, Wang M, Yu L, Wang J, Shen L, He K.
\newblock {Deep learning for digitizing highly noisy paper-based ECG records}.
\newblock Computers in Biology and Medicine 2020;\hspace{0pt}127:104077.

\end{thebibliography}

%%%%%%%%%%%%%%%%%%%%%%%%%%%%%%%%%%%%%%%%%%%%%%%%%
\begin{correspondence}
Felix Krones\\
1 St Giles, Oxford, OX1 3JS, UK\\
felix.krones@oii.ox.ac.uk
\end{correspondence}
%%%%%%%%%%%%%%%%%%%%%%%%%%%%%%%%%%%%%%%%%%%%%%%%%

\end{document}